\documentclass{article} 
\usepackage{iclr2016_conference,times}
\usepackage{hyperref}
\usepackage{url}
\usepackage{times}
\usepackage{epsfig}
\usepackage{graphicx}
\usepackage{amsmath}
\usepackage{amssymb}
\usepackage{multirow}
\usepackage{caption}

\newcommand{\bmx}[0]{\begin{bmatrix}}
\newcommand{\emx}[0]{\end{bmatrix}}

\newcommand{\vect}[1]{\mathbf{#1}}
\newcommand{\vects}[1]{\boldsymbol{#1}}

\newcommand{\vx}[0]{\vect{x}}

\newcommand{\TT}[0]{\vects{\theta}}

\title{Delving Deeper into Convolutional Networks for Learning Video Representations}

\author{
Nicolas Ballas$^1$, Li Yao$^1$, Chris Pal$^2$, Aaron Courville$^1$\\
$^1$MILA, Universit\'e de Montr\'eal.\\
$^2$\'Ecole Polytechnique de Mont\'real.\\
}
\iclrfinalcopy 

\begin{document}

\maketitle

\begin{abstract}

We propose an approach to learn spatio-temporal features in
videos from intermediate visual representations we call ``percepts'' using
Gated-Recurrent-Unit Recurrent Networks (GRUs).
Our method relies on percepts that are extracted from all levels of
a deep convolutional network trained on the large ImageNet dataset.
While high-level percepts contain highly discriminative information, 
they tend to have a low-spatial resolution.  
Low-level percepts, on the other hand, 
preserve a higher spatial resolution from which we can model finer motion patterns.

Using low-level percepts, however, can lead to high-dimensionality video representations. 
To mitigate this effect and control the number of parameters, 
we introduce a variant of the GRU model that leverages the convolution operations to enforce
sparse connectivity of the model units and share parameters across the
input spatial locations.

We empirically validate our approach on both Human Action Recognition and Video Captioning tasks.
In particular, we achieve results equivalent to state-of-art on the YouTube2Text dataset
using a simpler caption-decoder model and without extra 3D CNN features.

\end{abstract}

\section{Introduction}
Video analysis and understanding represents a major challenge for computer vision and machine learning research.
While previous work has traditionally relied on  hand-crafted and
task-specific  representations~\citep{wang.2011, sadanand2012action}, there
is a growing interest in designing general video representations that
could help solve tasks in video understanding such as human action recognition, video retrieval or video captionning~\citep{tran2014c3d}.

Two-dimensional Convolutional Neural Networks (CNN) have exhibited  state-of-art performance in still image tasks such as classification or detection~\citep{simonyan2014very}. However,  such models discard temporal information that has been shown to provide important cues in videos~\citep{wang.2011}.
On the other hand, recurrent neural networks (RNN) have demonstrated the ability to understand temporal sequences in various learning tasks such as speech recognition~\citep{graves2014towards} or machine translation~\citep{bahdanau2014neural}.
Consequently, Recurrent Convolution Networks (RCN) \citep{srivastava2015unsupervised, donahue2014long, ng2015beyond} that leverage both recurrence and convolution have recently been introduced for learning video representation.
Such approaches typically extract ``visual percepts'' by applying  a 2D CNN on the video frames and then feed the CNN activations to an RNN
in order to characterize the video temporal variation.

\begin{figure}
\center
\includegraphics[width=10cm]{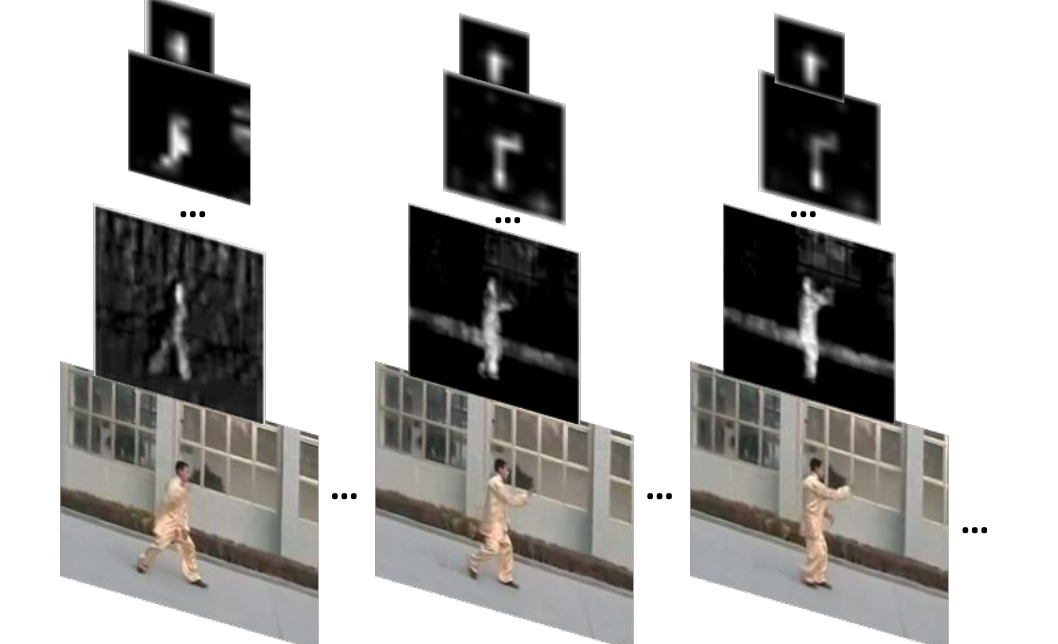}
\caption{Visualization of convolutional maps on successive frames
in video. As we go up in the CNN hierarchy, we observe that the convolutional maps are more stable over time, and thus discard variation over short temporal windows.}
\label{fig:motiv}
\end{figure}

Previous works on RCNs has tended to focus on high-level visual percepts extracted from the 2D CNN top-layers.
CNNs, however, hierarchically build-up spatial invariance through pooling 
layers~\citep{lecun1998gradient,simonyan2014very} as Figure~\ref{fig:motiv} highlights.
While CNNs tends to discard local information in their top layers, 
frame-to-frame temporal variation is known to be smooth. The motion of video patches tend to be restricted to a local neighborhood~\citep{brox2011large}.
For this reason, we argue that current RCN architectures are not well suited for capturing
fine motion information. Instead, they are more likely focus on global
appearance changes such as shot transitions.
To address this issue, we introduce a novel RCN architecture that applies an RNN not solely 
on the 2D CNN top-layer but also on the intermediate convolutional layers.
Convolutional layer activations, or convolutional maps, preserve a finer spatial resolution of 
the input video from which local spatio-temporal patterns are extracted.

Applying an RNN directly on intermediate convolutional maps, however, inevitably results in a drastic number
of parameters characterizing the input-to-hidden transformation due to the convolutional maps size.
On the other hand, convolutional maps preserve the frame spatial topology. 
We propose to leverage this topology by introducing sparsity and locality in the RNN units to reduce 
the memory requirement. We extend the GRU-RNN model~\citep{cho2014learning} and replace the fully-connected 
RNN linear product operation with a convolution. Our GRU-extension therefore
encodes the locality and temporal smoothness prior of videos directly in the model structure.

We evaluate our solution on UCF101 human action recognition from \citet{soomro2012ucf101} as well as the YouTube2text video captioning dataset from \citet{chen2011collecting}.
Our experiments show that leveraging ``percepts'' at multiple resolutions to model temporal variation
improves performance over our baseline model with respective gains of $3.4\%$  for action recognition and $10\%$ for video captioning.

\section{GRU: Gated Recurrent Unit Networks}

In this section, we review Gated-Recurrent-Unit (GRU) networks which
are a particular type of RNN.
An RNN model is applied to a sequence of inputs, which can have variable lengths. It defines a recurrent hidden state whose activation at each time is dependent on that of the previous time.
Specifically, given a sequence
$\mathbf{X}=(\mathbf{x}_1, \mathbf{x}_2 , ... , \mathbf{x}_T )$, the RNN  hidden state at time $t$ is defined as
$\mathbf{h}_t = \phi(\mathbf{h}_{t-1}, \mathbf{x}_t)$,  where $\phi$  is a nonlinear activation function.
RNNs are known to be difficult to train due to the 
exploding or vanishing gradient effect~\citep{bengio1994learning}.
However, variants of RNNs such as Long Short Term Memory  (LSTM)~\citep{hochreiter1997long} or Gated Recurrent Units (GRU)~\citep{cho2014learning} have empirically demonstrated their ability to model long-term temporal dependency in various task such as machine translation or image/video caption generation. In this paper, we will mainly focus on GRU networks as they have shown similar performance to LSTMs but with a lower memory requirement~\citep{chung2014empirical}.


GRU networks allow each recurrent unit to adaptively capture dependencies of different time scales. The activation $\mathbf{h}_t$ of the GRU is defined by the following equations:
\begin{eqnarray}
\mathbf{z}_t & = &  \sigma(\mathbf{W}_z \mathbf{x}_t + \mathbf{U}_z \mathbf{h}_{t-1}),\\
\mathbf{r}_t & = & \sigma(\mathbf{W}_r \mathbf{x}_t + \mathbf{U}_r \mathbf{h}_{t-1}),\\
\mathbf{\tilde{h}}_t & = & \tanh(\mathbf{W} \mathbf{x}_t + \mathbf{U} (\mathbf{r}_t \odot \mathbf{h}_{t-1}),\\
\mathbf{h}_t & = &(1 - \mathbf{z}_t) \mathbf{h}_{t-1} +  \mathbf{z}_t  \mathbf{\tilde{h}}_t,
\end{eqnarray}
where $\odot$ is an element-wise multiplication.
$\mathbf{z}_t$ is an update gate that decides the degree to which the unit updates its activation, or content. $\mathbf{r}_t$ is a reset gate. $\sigma$ is the sigmoid function.
When a unit $r_t^i$ is close to 0, the reset gate forgets the previously computed state, 
and makes the unit act as if it is reading the first symbol of an input sequence.
$\mathbf{\tilde{h}}_t$ is a candidate activation which is computed similarly to that of the traditional recurrent unit in an RNN.


\section{Delving Deeper into Convolutional Neural Networks}


This section delves into the main contributions of this work.
We aim at leveraging visual percepts from different convolutional levels
in order to capture temporal patterns that occur at different spatial resolution.

Let's consider $(\mathbf{x}^1_t, ..., \mathbf{x}^{L-1}_t, \mathbf{x}^{L}_t)_{(t=1..T)}$, a set of 2D convolutional maps extracted from $L$ layers at different time steps in a video.
We propose two alternative RCN architectures, GRU-RCN, and Stacked-GRU-RCN (illustrated in Figure~\ref{fig:motiv}) that combines information extracted from those convolutional maps.
\begin{figure}
\center
\includegraphics[width=13cm]{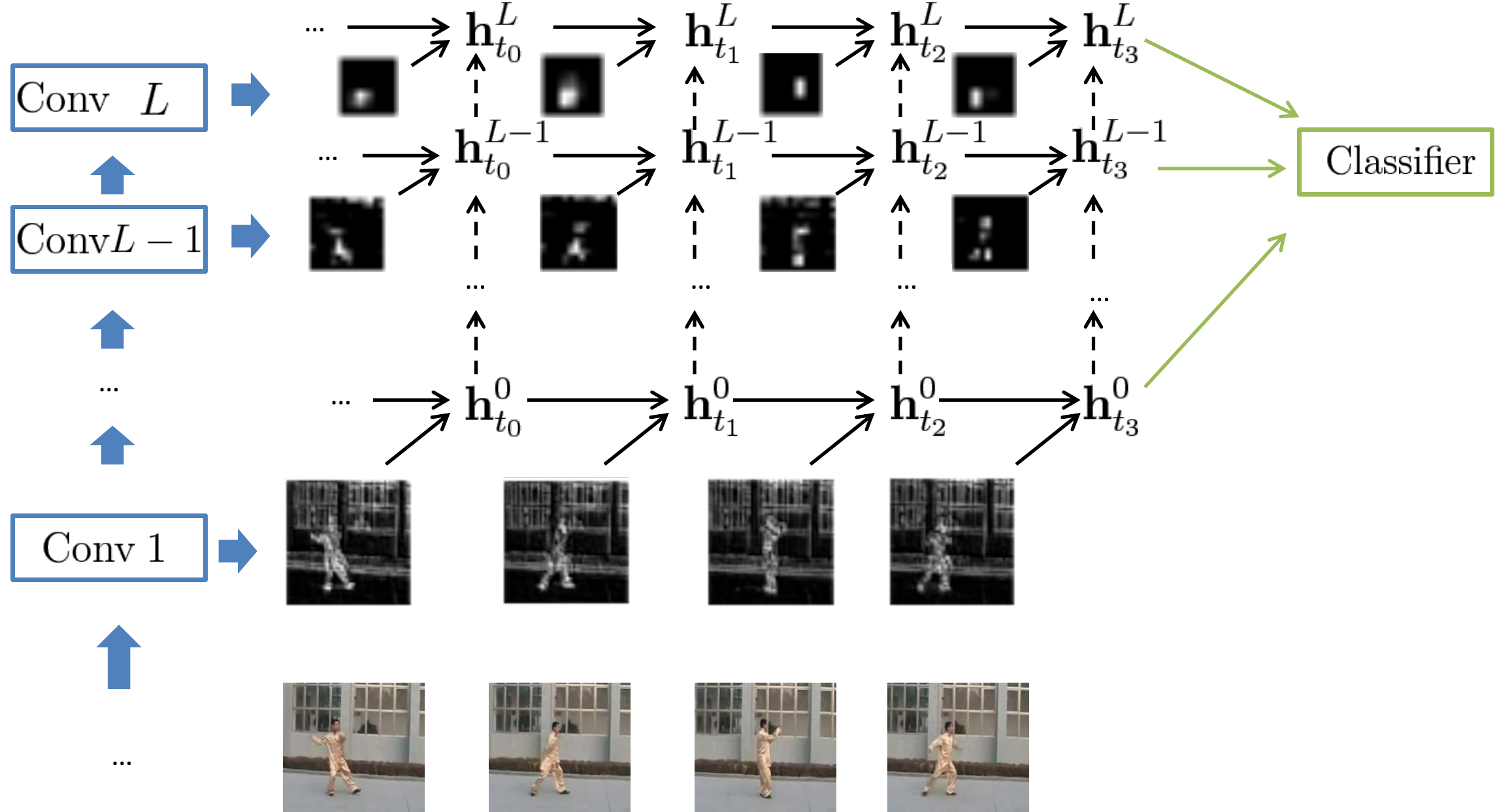}
\caption{High-level visualization of our model.
Our approach leverages convolutional maps from different layers
of a pretrained-convnet. Each map is given as input to a convolutional GRU-RNN (hence GRU-RCN) at different time-step. 
Bottom-up connections may be optionally added between RCN layers to form Stack-GRU-RCN.}
\label{fig:motiv}
\end{figure}
\subsection{GRU-RCN: }

In the first RCN architecture, we propose to apply $L$ RNNs independently on each convolutional map.
We define $L$ RNNs as $\phi^1, ...,\phi^L$, 
such that $\mathbf{h}^l_t = \phi^l(\mathbf{x}^l_t, \mathbf{h}^l_{t-1})$.
The hidden representation of the final time step $\mathbf{h}^1_T, ..., \mathbf{h}^L_T$ are then fed 
to a classification layer in the case of action recognition, or to a text-decoder RNN for caption generation.

To implement the RNN recurrent function $\phi^l$, we propose to leverage Gated Recurrent 
Units~\citep{cho2014learning}.
GRUs were originally introduced for machine translation. They model input to hidden-state and hidden to hidden transitions using fully connected units.
However, convolutional map inputs are 3D tensors (spatial dimension and input channels). Applying a GRU directly can lead to a drastic number of parameters.
Let $N_1$, $N_2$ and $O_x$ be the input convolutional map spatial size and number of channels. Applying a GRU directly would require input-to-hidden parameters $\mathbf{W}^l$,  $\mathbf{W}^l_z$ and $\mathbf{W}^l_r$ to be of size $N_1 \times N_2 \times O_x \times O_h$ where $O_h$ is the dimensionality of the GRU hidden representation.

Fully-connected GRUs do not take advantage of the underlying structure of convolutional maps.
Indeed, convolutional maps are extracted from images that are composed of patterns with 
strong local correlation which are repeated over different spatial locations.
In addition, videos have smooth temporal variation over time,
\textit{i.e.} motion associated with a given patch in successive frames will be  restricted in a local spatial neighborhood.
We embed such a prior in our model structure and replace
the fully-connected units in GRU with convolution operations. We therefore obtain
recurrent units that have sparse connectivity and share their parameters across different 
input spatial locations:
\begin{eqnarray}
\mathbf{z}^l_t & = & \sigma(\mathbf{W}^l_z * \mathbf{x}^l_t + \mathbf{U}^l_z * \mathbf{h}^l_{t-1}),\\
\mathbf{r}^l_t & = & \sigma(\mathbf{W}^l_r * \mathbf{x}^l_t + \mathbf{U}^l_r * \mathbf{h}^l_{t-1}),\\
\mathbf{\tilde{h}}^l_t &  = & \tanh(\mathbf{W}^l * \mathbf{x}^l_t + \mathbf{U} * (\mathbf{r}^l_t \odot \mathbf{h}^l_{t-1}),\\
\mathbf{h}^l_t &  = & (1 - \mathbf{z}^l_t) \mathbf{h}^l_{t-1} +  \mathbf{z}^l_t  \mathbf{\tilde{h}}^l_t,
\end{eqnarray}
where $*$ denotes a convolution operation.
In this formulation,  Model parameters $\mathbf{W}, \mathbf{W}^l_z, \mathbf{W}^l_r$ and $\mathbf{U}^l, \mathbf{U}^l_z, \mathbf{U}^l_r$ are 2D-convolutional kernels.
Our model results in hidden recurrent representation that preserves the
spatial topology, $\mathbf{h}^l_t = (\mathbf{h}^{l}_{t}(i, j))$
 where $\mathbf{h}^{l}_{t}(i, j))$ is a feature vector defined at the location $(i, j)$.
To ensure that the spatial size of the hidden representation remains fixed over time, we use zero-padding in the recurrent convolutions.

Using convolution, parameters $\mathbf{W}^l$,  $\mathbf{W}^l_z$ and $\mathbf{W}^l_r$ have a size of $k_1\times k_2 \times O_x \times O_h$ where $k_1\times  k_2$
is the convolutional kernel spatial size (usually $3\times 3$), chosen to be  significantly lower than convolutional map size $N_1\times N_2$.
The candidate hidden representation $\mathbf{\tilde{h}}_t(i,j)$, the activation gate $\mathbf{z}_k(i,j)$ and the reset gate $\mathbf{r}_k(i,j)$ are defined based on a local neigborhood of size $(k_1 \times k_2)$ at the location$(i, j)$ in both the input data $\mathbf{x}_t$ and the previous hidden-state $\mathbf{h}_{t-1}$.
In addition, the size of receptive field associated with $\mathbf{h}^l(i,j)_t$ increases in the previous presentation $\mathbf{h}^l_{t-1}, \mathbf{h}^l_{t-2}...$ as we go back further in time.
Our model is therefore capable of characterizing spatio-temporal patterns with high spatial variation in time.

A GRU-RCN layer applies 6 2D-convolutions at each time-step (2 per GRU gate and 2 for computing the candidate activation).
If we assume for simplicity that the input-to-hidden and hidden-to-hidden convolutions have the same
kernel size and perserve the input dimension, GRU-RCN requires $O(3T N_1 N_2 k_1 k_2 (O_x O_h + O_h O_h))$ multiplications.
GRU-RCN sparse connectivity therefore saves computation compared to a fully-connected RNN that would require $O(3T N_1 N_2 N_1 N_2(O_x O_h + O_h O_h))$ computations.
Memorywise, GRU-RCN needs to store the parameters for all 6 convolutions kernels leading to $O(3 k_1 k_2 (O_x O_h + O_h O_h))$ parameters.


\subsection{Stacked GRU-RCN:}

In the second RCN architecture, we investigate the importance of bottom-up connection across RNNs.
While GRU-RCN applies each layer-wise GRU-RNN in an independent fashion,
Stacked GRU-RCN preconditions each GRU-RNN on the output of the previous GRU-RNN at the current time step: $\mathbf{h}^l_t = \phi^l(\mathbf{h}^l_{t-1}, \mathbf{h}^{l-1}_{t}, \mathbf{x}^l_t)$.
The previous RNN hidden representation is given as an extra-input to the GRU convolutional units:
\begin{eqnarray}
\mathbf{z}^l_l & = & \sigma(\mathbf{W}^l_z * \mathbf{x}^l_t + \mathbf{W}^l_{z^l} * \mathbf{h}^{l-1}_t + \mathbf{U}^l_z * \mathbf{h}^l_{t-1}),\\
\mathbf{r}^l_t & = &  \sigma(\mathbf{W}^l_r * \mathbf{x}^l_t + \mathbf{W}^l_{r^l} * \mathbf{h}^{l-1}_t  + \mathbf{U}^l_r \mathbf{h}^l_{t-1}),\\
\mathbf{\tilde{h}}^l_t &  = & \tanh(\mathbf{W}^l * \mathbf{x}^l_t + \mathbf{U}^l * (\mathbf{r}_t \odot \mathbf{h}^l_{t-1}),\\
\mathbf{h}^l_t &  = & (1 - \mathbf{z}^l_t) \mathbf{h}^l_{t-1} +  \mathbf{z}^l_t  \mathbf{\tilde{h}}^l_t,
\end{eqnarray}
Adding this extra-connection brings more flexibility and gives the opportunity for the model to leverage representations with different resolutions.


\section{Related Work}
Deep learning approaches have recently been used to learn video representations
and have produced state-of-art results~\citep{karpathy2014large, simonyan2014two, wang2015towards, tran2014c3d}.
~\citet{karpathy2014large, tran2014c3d} proposed  to  use 3D CNN learn a video representations,
 leveraging large training datasets such as the Sport 1 Million.
However, unlike image classification~\citep{simonyan2014very},  CNNs did not yield large improvement over these traditional methods~\citep{lan2014beyond} highlighting the difficulty of learning video representations even with large training dataset.
\citet{simonyan2014two} introduced a two-stream framework where they train CNNs independently on  RGB and optical flow inputs. While the flow stream focuses only on motion information, the RGB stream can leverage 2D CNN pre-trained on image datasets.
Based on the Two Stream representation, ~\cite{wang2015action} extracted deep feature and conducted trajectory constrained pooling to aggregate convolutional feature as video representations.

RNN models have also been used to encode temporal information for learning video representations in conjonction with 2D CNNs.
\cite{ng2015beyond,donahue2014long} applied an RNN on top of the the two-stream framework, while
\cite{srivastava2015unsupervised} proposed, in addition, to investigate the benefit of learning a video representation in an unsupervised manner.
Previous works on this topic has tended to focus only on high-level CNN ``visual percepts''. In contrast, our approach proposes to leverage visual ``percepts'' extracted from different layers in the 2D-CNN.

Recently, \cite{shi2015convolutional} also proposed to leverage convolutional units inside an RNN network.
However, they focus on different task (now-casting) and a different RNN model based on an LSTM.
In addition, they applied their model directly on pixels. Here, we use
recurrent convolutional units on pre-trained CNN convolutional maps,
to extract temporal pattern from  visual ``percepts'' with different spatial sizes.

\section{Experimentation}
This section presents an empirical evaluation of the proposed GRU-RCN and Stacked GRU-RCN architectures.
We conduct experimentations on two different tasks: human action recognition and video caption generation.

\subsection{Action Recognition}
We evaluate our approach on the UCF101 dataset~\cite{soomro2012ucf101}.
This dataset  has 101 action classes spanning over 13320  YouTube  videos  clips.
Videos composing the dataset are subject to large camera motion,  viewpoint
change  and  cluttered  backgrounds.
We report results on the dataset UCF101 first split, as this is most commonly used split in the literature.
To perform proper hyperparameter seach, we use the videos from the UCF-Thumos
validation split~\cite{jiang2014thumos} as the validation set.




\subsubsection{Model Architecture}
In this experiment, we consider the RGB and flow representations of videos as  inputs.
We extract visual ``percept'' using VGG-16 CNNs that consider either RGB or flow inputs. VGG-16 CNNs are pretrained on ImageNet~\citep{simonyan2014very} and fine-tuned on the UCF-101 dataset, following the protocol in \citet{wang2015towards}.
We then extract the convolution maps from \textit{pool2}, \textit{pool3}, \textit{pool4}, 
\textit{pool5} layers and the fully-connected map from layer \textit{fc-7} 
(which can be view as a feature map with a $1\times 1$ spatial
dimension). Those features maps are given as inputs to our RCN models.

We design and evaluate three RCN architectures for action recognition.
In the first RCN architecture, GRU-RCN, we apply 5 convolutional GRU-RNNs independently
on each convolutional map.
Each convolution in the GRU-RCN has zero-padded $3\times 3$ convolutions that 
preserves the spatial dimension of the inputs . 
The number of channels of each respective GRU-RNN hidden-representations 
are $64$, $128$, $256$, $256$, $512$.
After the RCN operation we obtain $5$ hidden-representations for each time step.
We apply average pooling on the hidden-representations of the last time-step
to reduce their spatial dimension to $1\times 1$, and feed the representations to $5$ classifiers, composed by a linear layer with a softmax nonlineary.
Each classifier therefore focuses on only 1 hidden-representation extracted from
the convolutional map of a specific layer.
The classifier outputs are then averaged to get the final decision. 
A dropout ratio of $0.7$ is applied  on the input of each classifiers.

In the second RCN architecture, Stacked GRU-RCN, we investigate the usefulness of bottom-up connections.
Our stacked GRU-RCN uses the same base architecture as the GRU-RCN, consisting of 
5 convolutional GRU-RNNs having $64, 128, 256, 256$ channels respectively.
However, each convolutional GRU-RNN is now preconditioned on the hidden-representation that the GRU-RNN applied on the previous convolution-map outputs.
We apply max-pooling on the hidden representations between the GRU-RNN layers for the compatibility
of the spatial dimensions.
As for the previous architecture, each GRU-RNN hidden-representation at the
last time step is pooled and then given as input to a classifier.

Finally, in our bi-directional GRU-RCN, we investigate the importance of reverse temporal information.
Given convolutional maps extracted from one layer,
we run the GRU-RCN twice, considering the inputs in both sequential and reverse temporal order. 
We then concatenate the last hidden-representations of the foward GRU-RCN and backward GRU-RCN, and give the resulting vector to a classifier.
\subsubsection{Model Training and Evaluation}

We follow the training procedure introduced by the two-stream framework~\cite{simonyan2014two}.
At each iteration, a batch of 64 videos are sampled randomly from the the training set.
To perform scale-augmentation, we randomly sample the cropping width and height from $256,224,192,168$.
The temporal cropping size is set to $10$. We then resize the cropped volume to $224\times 224\times 10$.
We estimate each model parameters by maximizing the model log-likelihood:
$$
 \mathcal{L}(\TT) = \frac{1}{N} \sum_{n=1}^N \log p(y^n \mid c(\vx^n), \TT),
$$
 where there are $N$ training video-action pairs $(\vx^n, y^n)$, $c$ is a function that takes a crop
 at random.
We use Adam~\cite{kingma2014adam} with the gradient computed by the
backpropagation algorithm. We perform early stopping and choose the parameters
that maximize the log-probability of the validation set.

We also follow the evaluation protocol of the two-stream framework~\cite{simonyan2014two}.
At the test time, we sample 25 equally spaced video sub-volumes with a temporal size of 10 frames.
From each of these selected sub-volumes, we obtain
10 inputs for our model, i.e.  4 corners, 1 center, and their horizontal flipping.  The final pre-
diction score is obtained by averaging across the sampled sub-volumes and their cropped regions.



\subsubsection{Results}
\begin{table}[h]
\center
\begin{tabular}{c|c|c}
Method & RGB & Flow\\
\hline
VGG-16      & 78.0 & 85.4\\
VGG-16 RNN  & 78.1 & 84.9\\
\hline
GRU-RCN     &  79.9 & \textbf{85.7}\\
Stacked-GRU RCN  & 78.3& -\\
Bi-directional GRU-RCN  & \textbf{80.7} &-\\
\hline
Two-Stream{\small~\cite{simonyan2014very}} & 72.8 & 81.2\\
Two-Stream + LSTM{\small~\cite{donahue2014long}}  & 71.1 & 76.9\\
Two-Stream + LSTM + Unsupervised{\small~\cite{srivastava2015unsupervised}}  & 77.7 & 83.7\\
Improved Two-Stream{\small~\cite{wang2015towards}} & \textbf{79.8} & \textbf{85.7}\\
\hline
 C3D one network{\small~\cite{tran2014c3d}, 1 million videos as training} & 82.3&-\\
 C3D ensemble{\small~\cite{tran2014c3d}, 1 million videos as training} & \textbf{85.2}&-\\
 Deep networks{\small~\cite{karpathy2014large}, 1 million videos as training} & 65.2&-\\

\end{tabular}
\caption{Classification accuracy of different variants of the model on the UCF101 split 1. We report the performance of previous works that learn representation using only RGB information only.}
\label{tab:ucf101}

\end{table}

We compare our approach with two different baselines, VGG-16 and VGG-16 RNN.
VGG-16 is the 2D spatial stream that is described in~\cite{wang2015towards}.
We take the VGG-16 model, pretrained on Image-Net and fine-tune it on the UCF-101 dataset.
VGG-16 RNN baseline applied an RNN, using fully-connected gated-recurrent units, on top-of VGG-16.
It takes as input the VGG-16 fully-connected representation \textit{fc-7}.
Following GRU-RCN top-layer, the VGG-16 RNN has hidden-representation dimensionality of $512$.

The first column of Table~\ref{tab:ucf101} focuses on RGB inputs.
We first report results of different GRU-RCN variants
and compare them with the two baselines: VGG-16 and VGG-16 RNN.
Our GRU-RCN variants all outperform the baselines, showing the benefit of delving deeper into a CNN in order to learn a video representation. We notice that VGG-16 RNN only slightly improve over the VGG-16 baseline, $78.1$ against $78.0$. This result confirms that CNN top-layer tends to discard temporal variation over short temporal windows.
Stacked-GRU RCN performs significantly lower than GRU-RCN and Bi-directional GRU-RCN. We argue that bottom-up connection, increasing the depth of the model, combined with the lack of training data (UCF-101 is train set composed by only ~9500 videos) make the Stacked-GRU RCN learning difficult.
The bi-directional GRU-RCN performs the best among the GRU-RCN variant with an accuracy of $80.7$, showing the advantage of modeling temporal information in both sequential and reverse order.
Bi-directional GRU-RCN obtains a gain $3.4\%$ in term of performances, relatively to the baselines that focus only the VGG-16 top layer.

Table~\ref{tab:ucf101} also reports results from other state-of-art approaches using RGB inputs.
C3D~\cite{tran2014c3d} obtains the best performance on UCF-101 with $85.2$. However, it should be noted that C3D is trained over 1 million videos. Other approaches use  only the 9500 videos of UCF101 training set for learning temporal pattern.
Our Bi-directional GRU-RCN compare favorably with other Recurrent Convolution Network (second blocks), confirming
the benefit of using different CNN layers to model temporal variation.

Table~\ref{tab:ucf101} also evaluates the GRU-RCN model applied flow inputs. VGG-16 RNN baseline actually decreases the performance compared to the VGG-16 baseline. On the other hand, GRU-RCN outperforms the VGG-16 baseline achieving $85.7$ against $85.4$. While the improvement is less important than the RGB stream, it should be noted that the flow stream of VGG-16 is applied on 10 consecutive flow inputs to extract visual ``percepts'', and therefore already captures some motion information.

Finally, we investigate the combination of the RGB and flow streams.
Following~\cite{wang2015towards}, we  use a weighted linear combination
of their prediction scores, where the weight is set to $2$ as for the flow stream net and $1$ for the temporal stream. Fusion
the VGG-16 model baseline achieve an accuracy of $89.1$. Combining the RGB Bi-directional GRU-RCN with the flow GRU-RCN achieves
a performance gain of $1.9\%$ over baseline, reaching $90.8$. Our model is on part with~\cite{wang2015towards} that obtain  state-of-art results using both RGB and flow streams which obtains $90.9$.

\subsection{Video Captioning}
We also  evaluate our representation on the video captioning task using YouTube2Text video corpus~\cite{chen2011collecting}.
The  dataset  has  1,970  video clips with multiple natural language descriptions for each
video clip.   The dataset is open-domain and covers a wide range of topics such as sports,
animals, music and movie clips. Following~\cite{yao2015describing}, we split the dataset into a training set of 1,200 video clips, a validation set of 100
clips and a test set consisting of the remaining clips.

\subsubsection{Model Specifications}

To perform video captioning, we use the so-called encoder-decoder framework~\cite{cho2014learning}.
In this framework the encoder maps input videos into abstract representations that precondition a caption-generating decoder.

As for encoder, we compare both VGG-16 CNN and Bi-directional GRU-RCN. 
Both models have been fine-tuned on the UCF-101 dataset
and therefore focus on detecting \emph{actions}.
To extract an abstract representation from a video, we sample $K$ equally-space segments.
When using the VGG-16 encoder, we provide the $fc7$ layer activations of the  each segment's first frame as the input to the text-decoder. For the GRU-RCN, we apply our model on the  segment's 10 first frames. We concatenate the GRU-RCN hidden-representation from the last time step. The concatenated vector is given as 
the input to the text decoder.
As it has been shown that characterizing \emph{entities} in addition of \emph{action} is important for the
 caption-generation task~\cite{yaoliupperbound}, we also use as encoder a CNN~\cite{szegedy2014going},
pretrained on ImageNet, that focuses on detecting static visual object categories.

As for the decoder, we use an LSTM text-generator with soft-attention on the 
video temporal frames~\cite{yao2015describing}.

\subsubsection{Training}

For all video captioning models, we estimated the parameters $\TT$
of the decoder by maximizing the log-likelihood:
\begin{align*}
    \mathcal{L}(\TT) = \frac{1}{N} \sum_{n=1}^N \sum_{i=1}^{t_n} \log p(y_i^n \mid
    y_{<i}^n, \vx^n, \TT),
\end{align*}
where there are $N$ training video-description pairs $(\vx^n, y^n)$, and each
description $y^n$ is $t_n$ words long. 
We used Adadelta~\cite{Zeiler-2012} We optimized the hyperparameters (e.g. number of LSTM units and the word
embedding dimensionality, number of segment $K$) using random search \citep{bergstra2012random} 
to maximize the log-probability of the validation set.

\subsubsection{Results}

\begin{table*}
\centering
\begin{tabular}{l||c|ccc|}
 &\multicolumn{4}{c|}{YouTube2Text} \\
\hline
  Model                        &model selection &{\small BLEU} & {\small METEOR} & {\small CIDEr}  \\
\hline
VGG-16 Encoder     &  BLEU  &0.3700  & 0.2640 & 0.4330\\
Bi-directional GRU-RCN Encoder    &  BLEU  &0.4100   & 0.2850  & 0.5010\\
GoogleNet Encoder  &  BLEU  &0.4128   & 0.2900   & 0.4804\\
GoogleNet + Bi-directional GRU-RCN Encoder & BLEU   &\bf 0.4963   & 0.3075   & 0.5937\\
GoogleNet + Bi-directional GRU-RCN Encoder & NLL   &0.4790   & 0.3114   & \bf 0.6782\\
GoogleNet + Bi-directional GRU-RCN Encoder & METEOR   &0.4842   & \bf 0.3170   & 0.6538\\
GoogleNet + Bi-directional GRU-RCN Encoder & CIDEr   &0.4326   & \bf 0.3160   & \bf 0.6801\\
\hline
GoogleNet + HRNE \citep{pan2015hierarchical}      &-& 0.436       & \bf 0.321 & -\\
VGG + p-RNN \citep{Haonan2015}                          &-& 0.443       & 0.311 & -\\
VGG + C3D + p-RNN \citep{Haonan2015}                          &-& \bf 0.499       & \bf 0.326 & -\\
Soft-attention~\cite{yao2015describing}     &-&  0.4192     &  0.2960 &  0.5167\\
Venugopalan~\textit{et al.}~\cite{venugopalan2014translating} &-& 0.3119     & 0.2687 & -     \\
+ Extra Data (Flickr30k, COCO)                                 &-& 0.3329     & 0.2907 & -     \\
\hline
Thomason ~\textit{et al.}~\cite{thomason2014} &-&0.1368     & 0.2390 & -            \\ 
 \end{tabular}
\caption{Performance of different variants of the model on YouTube2Text for video captioning. Representations obtained with the proposed RCN architecture combined with decoders from \citet{yao2015describing} offer a significant performance boost, reaching the 
performance of the other state-of-the-art models.}
\label{tab:y2t}
 \end{table*}

Table~\ref{tab:y2t} reports the performance of our proposed method using three automatic evaluation metrics. 
These are BLEU in \citet{papineni2002bleu}, METEOR in \citet{meteor:2014}
and CIDEr in ~\citet{vedantam2014cider}. 
We use the evaluation script prepared and introduced in \citet{chen2015microsoft}.
All models are early-stopped based on the negative-log-likelihood (NLL) of the validation set.
We then select the model that performs best on the validation set according to the metric at consideration.

The first two lines of Table~\ref{tab:y2t} compare the performances of the VGG-16 and Bi-directional GRU-RCN encoder.
Results clearly show the superiority of the Bi-Directional GRU-RCN Encoder
as it outperforms the VGG-16 Encoder on all three metrics.
In particular, GRU-RCN Encoder obtains a performance gain of $10\%$ compared to the VGG-16 Encoder according to the BLEU metric.
Combining our GRU-RCN Encoder that focuses on \emph{action} with a GoogleNet Encoder that 
captures visual \emph{entities} further improve the performances.

Our GoogleNet + Bi-directional GRU-RCN approach significantly outperforms 
Soft-attention~\cite{yao2015describing} that relies on a GoogLeNet and cuboids-based 3D-CNN Encoder, 
in conjunction to a similar soft-attention decoder. This result indicates that our approach is able to 
offer more effective representations.
According to the BLEU metric, we also outperform other approaches using more complex decoder schemes such as spatial and temporal attention decoder~\citep{Haonan2015} or a hierarchical RNN decoder~\citep{pan2015hierarchical}
Our approach is on par with~\cite{Haonan2015}, without the need of using a C3D-encoder
that requires training on large-scale video dataset.

\section{Conclusion}

In this work, we address the challenging problem of learning discriminative and abstract representations from videos.
We identify and underscore the importance of modeling temporal
variation from ``visual percepts'' at  different spatial resolutions.
While high-level percepts contain highly discriminative information, they tend to have a low-spatial resolution.  Low-level percepts, on the other hand, preserve a higher spatial resolution from which we can model finer motion patterns.
We introduce a novel recurrent convolutional network architecture
that leverages convolutional maps, from all levels of
a deep convolutional network trained on the ImageNet dataset, to take advantage of ``percepts'' from different spatial resolutions.

We have empirically validated our approach on the Human Action Recognition and Video Captioning tasks using the UCF-101 and YouTube2Text datasets.
Experiments demonstrate that leveraging ``percepts'' at multiple resolutions to model temporal variation
improve over our baseline model, with respective gain of $3.4\%$ and $10\%$ for the action recognition  and video captions tasks using RGB inputs. In particular, we achieve results comparable to state-of-art on YouTube2Text
using a simpler text-decoder model and without extra 3D CNN features.

\section*{Acknowledgments}

The authors would like to acknowledge the support of the following agencies for
research funding and computing support: NSERC, Calcul Qu\'{e}bec, Compute Canada,
the Canada Research Chairs and CIFAR. We
would also like to thank the developers of Theano \citep{bergstra+al:2010-scipy,Bastien-Theano-2012}
, for developing such a
powerful tool for scientific computing.

\bibliography{iclr2016_conference}
\bibliographystyle{iclr2016_conference}

\end{document}